\documentclass[runningheads]{llncs}
\usepackage[T1]{fontenc}
\usepackage{graphicx}
\usepackage{mathptmx}
\usepackage{xcolor}
\mathchardef\mhyphen="2D
\usepackage{cite}
\usepackage{algorithm}
\usepackage{algorithmic}
\usepackage{amssymb}
\usepackage{xspace}
\usepackage{amsmath} 
\usepackage{xspace}
\newcommand{\eg}{e.\,g.,\xspace}
\newcommand{\ie}{i.\,e.,\xspace}
\newcommand{\cf}{cf.\xspace}

\newcommand{\PHeatPruner}{\textsc{PHeatPruner}\xspace} 

\usepackage{paralist}

\begin{document}
\title{\PHeatPruner: Interpretable Data-centric Feature Selection for Multivariate Time Series Classification through Persistent Homology}
\titlerunning{\PHeatPruner: Interpretable Feature Selection via Persistent Homology}
%
\author{Anh-Duy Pham\inst{1,5}\orcidID{0000-0003-3832-9453} \and \\
Olivier Basole Kashongwe\inst{1,5}\orcidID{0000-0002-6107-4084} \and \\
Martin Atzmueller \inst{1,3,4}\orcidID{0000-0002-2480-6901} \and
Tim R{\"o}mer\inst{1,2} 
}
\authorrunning{Pham et al.}
%
\institute{Osnabr{\"u}ck University, Joint Lab Artificial Intelligence \& Data Science, Osnabr{\"u}ck, Germany \and
Osnabr{\"u}ck University, Institute of Mathematics, Osnabr{\"u}ck, Germany
 \and
Osnabr{\"u}ck University, Semantic Information Systems Group, Osnabr{\"u}ck, Germany \and
German Research Center for Artificial Intelligence (DFKI),\\ Research Department Plan-Based Robot Control, Osnabr{\"u}ck, Germany \and
Leibniz Institute for Agricultural Engineering and Bioeconomy (ATB),\\ Department of Sensors and Modelling, Potsdam, Germany
\email{\{anh-duy.pham,olivier.kashongwe,martin.atzmueller,tim.roemer\}@uos.de}}
\maketitle              
\begin{abstract}
Balancing performance and interpretability in multivariate time series classification is a significant challenge due to data complexity and high dimensionality. This paper introduces \PHeatPruner, a method integrating persistent homology and sheaf theory to address these challenges. Persistent homology facilitates the pruning of up to $45\%$ of the applied variables while maintaining or enhancing the accuracy of models such as Random Forest, CatBoost, XGBoost, and LightGBM, all without depending on posterior probabilities or supervised optimization algorithms. Concurrently, sheaf theory contributes explanatory vectors that provide deeper insights into the data's structural nuances. The approach was validated using the UEA Archive and a mastitis detection dataset for dairy cows. The results demonstrate that \PHeatPruner effectively preserves model accuracy. Furthermore, our results highlight \PHeatPruner's key features, \ie simplifying complex data and offering actionable insights without increasing processing time or complexity. This method bridges the gap between complexity reduction and interpretability, suggesting promising applications in various fields.

\end{abstract}
\section{Introduction}
Effectively analyzing Multivariate Time Series (MTS) data is essential for decision-making in domains such as finance, healthcare, and environmental monitoring. Accurate MTS classification requires feature selection to identify key predictive variables, but traditional methods often struggle with the data's complexity and high dimensionality. MTS classifiers are categorized into similarity-based methods (\eg Dynamic Time Warping with k-NN)~\cite{shokoohi2017generalizing}, feature-based approaches, \eg shapelet models like Generalized Random Shapelet Forest (gRSF)~\cite{karlsson2016generalized} and Ultra-Fast Shapelets (UFS)~\cite{wistuba2015ultra}, Bag-of-Words models like WEASEL+MUSE)~\cite{schafer2017multivariate}, as well as deep learning methods, \eg LSTM, CNN, and Transformer architectures. Among these, the MLSTM-FCN model, which integrates LSTM and CNN layers with squeeze-and-excitation blocks, consistently shows superior performance~\cite{karim2019multivariate}.
Despite their effectiveness, these methods often lack interpretability, obscuring the features that drive their predictions. To tackle this, we employ persistent homology from Topological Data Analysis (TDA)~\cite{carlsson2021topological, edelsbrunner2014short}: it analyzes the shape of data in high dimensions by identifying significant features that persist across multiple scales. This approach is particularly valuable for MTS classification, as it captures the intrinsic topological structure of the data over time, enhancing model performance and interpretability by revealing how features evolve and interact.

Our contributions are threefold:
\begin{inparaenum}[(1)]
    \item We introduce a precise feature selection method: leveraging the persistent homology of the topological space formed by correlated features in MTS data, identifying features that are significant and stable over time for more accurate and robust classification; 
    \item We enhance this topological framework by incorporating sheaf theory~\cite{bredon2012sheaf}, enabling the generation of supplementary explanatory vectors through sheaf filtration from the simplices formed by connected features;
    \item Our evaluation using UEA Archive data as well as mastitis detection data demonstrates that our integrated approach maintains high levels of prediction accuracy and recall while providing deeper insights into the data structure, thereby improving overall interpretability.
\end{inparaenum}

\begin{figure}[h!]
\centering
\includegraphics[width=.95\columnwidth]{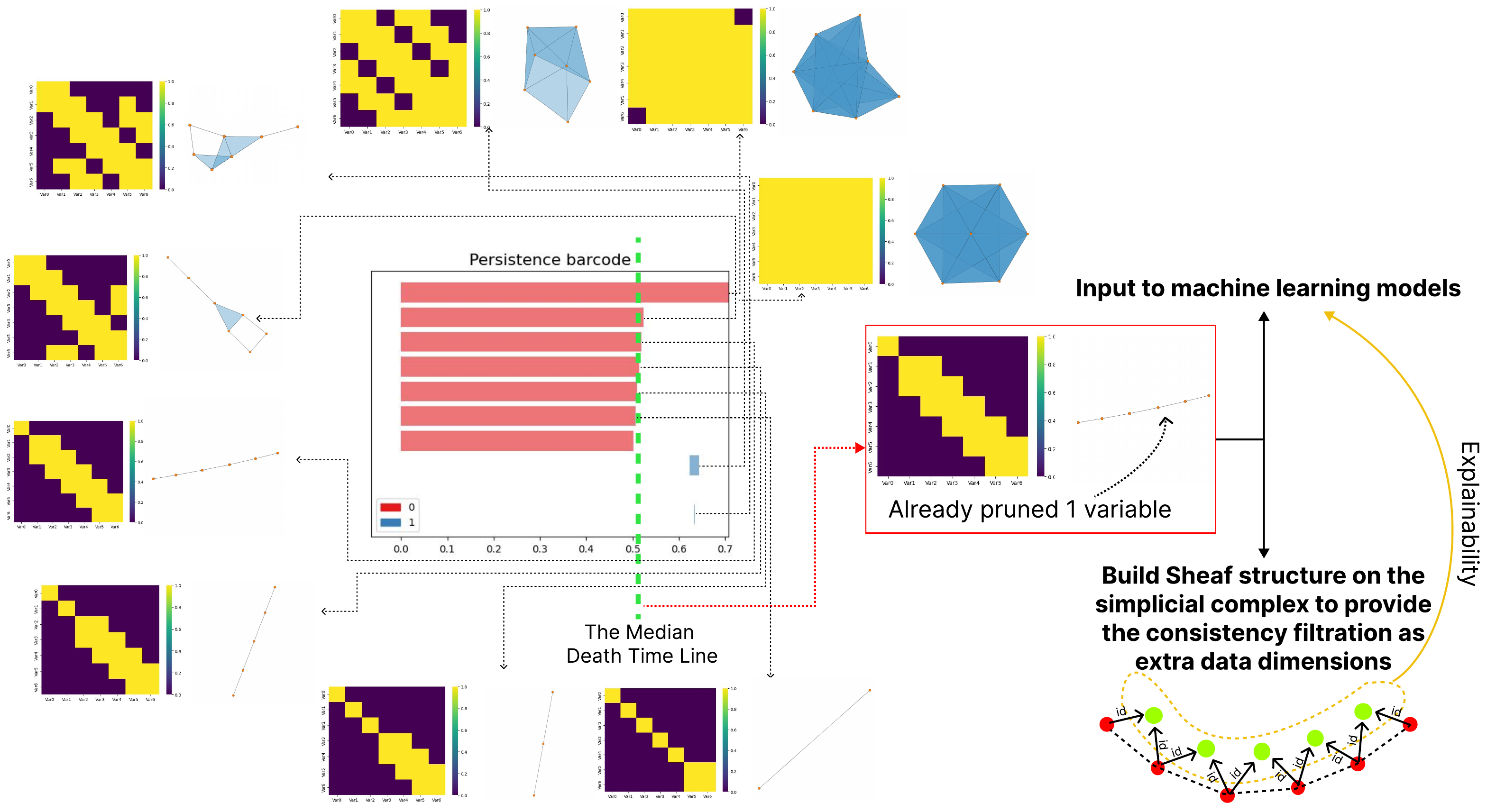}
\caption{Overview of \PHeatPruner.} \label{fig:overview}
\end{figure}

An overview on our proposed method is shown in Figure~\ref{fig:overview}. By integrating persistent homology and sheaf theory into MTS classification, we bridge the gap between model performance and interpretability. This approach not only advances our capability to handle complex, high-dimensional data but also provides clearer insights into the driving factors behind model predictions, facilitating more informed and actionable decision-making.

\section{Related Work}
Deep learning excels in MTS classification but often lacks interpretability. Attention-based models such as RETAIN~\cite{choi2016retain}, DA-RNN~\cite{qin2017dual}, DSTP-RNN~\cite{liu2020dstp}, GeoMAN~\cite{liang2018geoman}, and STAM~\cite{gangopadhyay2021spatiotemporal} improve focusing on relevant data points but still may not fully explain the overall model behavior. In order to enhance transparency, models like XCM~\cite{fauvel2021xcm} and TSEM~\cite{pham2023tsem} identify key features and integrate temporal and spatial aspects but struggle to quantify the impact of adding or removing features on classification outcomes. ShapeNet, introduced by Guozhong Li et al.~\cite{li2021shapenet}, integrates shapelets with neural networks to enhance the classification accuracy and interpretability of MTS data.

Multivariate feature selection methods for time series data are critical for improving model performance and interpretability. Traditional approaches like Principal Component Analysis (PCA) and Linear Discriminant Analysis (LDA) reduce dimensionality by capturing variance or maximizing class separation. Mutual Information-based and Rough Set-based methods evaluate dependencies between features and target variables to select significant features. Correlation-based techniques identify features with the highest correlation to the target, while wrapper and embedded methods incorporate feature selection into the model building process, often leading to high computational costs and model dependencies. Multi-Objective and Fuzzy Feature Selection methods introduce flexibility by optimizing multiple criteria and handling uncertainty, whereas Evolutionary and Random Probes-based methods explore feature subsets through adaptive algorithms~\cite{theng2024feature}. Although traditional feature selection methods are effective, they frequently rely on model outcomes for evaluation. This dependency can introduce biases and may not adequately capture complex, multi-scale interactions among features. In contrast, our approach uniquely integrates the optimization of feature selection with the generation of explanatory vectors, independent of model outcomes. 

Persistent homology from TDA~\cite{wasserman2018topological, chazal2021introduction} has been pivotal in examining time series data via sliding window embeddings, \cf Perea and Harer~\cite{perea2015sliding}. This method captures the topological features of time series across different scales~\cite{santoro2023higher} and has been applied across diverse domains, from gene expression analysis~\cite{dequeant2008comparison, perea2015sw1pers} to climate pattern investigation~\cite{berwald2013automatic}. Our approach enhances traditional methods by integrating the correlation matrix of the variables in the MTS through time as a distance function into the feature selection process, facilitating the identification of significant variables and filtering out less critical ones, thus streamlining the analysis of MTS data.

Sheaf theory~\cite{bredon2012sheaf, rosiak2022sheaf, curry2014sheaves, hansen2019toward, bodnar2023topological} offers a robust framework for integrating local datasets into a coherent global perspective~\cite{malcolm2009sheaves}. It can be used to organize heterogeneous sensor deployments~\cite{joslyn2014towards} and underpins data fusion algorithms~\cite{purvine2018topological}. Sheaf theory quantifies relationships through consistency radius and filtration, allowing detailed consistency analysis~\cite{robinson2017sheaves, robinson2020assignments}, and has applications in network analysis and quantum information~\cite{ghrist2011network, abramsky2015contextuality}. For example, sheaf theory supports a self-correcting algorithm for air quality monitoring~\cite{pham2023harnessing} and manages uncertainty in geolocation data~\cite{joslyn2020sheaf}, illustrating its capability to provide robust, topology-based solutions in complex data environments, thereby enhancing accuracy and reliability across various applications. \PHeatPruner leverages this advantage to provide a clear understanding of how each variable contributes to the prediction and how their relationships evolve over time.

\section{Persistent Homology Analysis of Correlation Matrices for Accurate Feature Extraction}
For a given MTS,  we compute the correlation matrix, representing relationships among variables. Interpreted as a weighted graph, this matrix is used in persistent homology to identify topological features like holes as the correlation threshold changes. These features are summarized in persistence diagrams or barcodes, offering a concise view of the data’s multi-scale relationships.

\subsection{Persistent Homology via Vietoris-Rips Complex}
Topological Data Analysis (TDA) employs, in particular, \textit{persistent homology} to analyze complex, multi-scale structures in datasets. This method examines the shape and connectivity of data by constructing a sequence of simplicial complexes from point clouds, defined by a resolution parameter \( \epsilon \). As \( \epsilon \) varies, these complexes evolve, and (topological) \textit{Betti numbers} \( \beta_i \) are computed to quantify topological features at different dimensions. Specifically, \( \beta_0 \) represents the number of connected components, \( \beta_1 \) indicates the number of loops, and \( \beta_2 \) identifies the number of voids. These classical Betti numbers together with \textit{persistent Betti numbers} collectively provide a concise summary of the data's multi-dimensional topological structure, revealing the underlying geometric and topological features across various scales.

\subsubsection{Vietoris-Rips Complex Construction from the Correlation Matrices}

In \PHeatPruner, for constructing Vietoris-Rips complexes from correlation matrices, we employ the \textit{Vietoris-Rips approach} (see Def. \ref{Def:VRC}), an important technique applied in TDA. This method constructs complexes by surrounding each data point with a ball of a designated radius $\epsilon$. A $k$-simplex is included in the complex if and only if the pairwise distances between all $k+1$ vertices are less than $\epsilon$. To apply this method to correlation matrices, we first transform the correlation matrix $C$, which quantifies the relationships between pairs of data points, into a \textit{distance matrix} $D$. This transformation is commonly performed using a function as simple as $D_{ij} = (1 - C_{ij})$. However, here we employ the distance function as follows:
\begin{equation}
D_{ij} = \sqrt{2(1 - C_{ij})}\,,
\end{equation}
to ensure that higher correlations are translated into shorter distances. For instance, when correlations are extremely high (\ie close to $1$), the linear transformation $1 - C_{ij}$ tends to compress the differences, making it challenging to differentiate between items that are very similar. In contrast, using the square root transformation $\sqrt{2(1 - C_{ij})}$ provides a better scale and differentiation, preserving proportional differences even among highly correlated items. Once the distance matrix $D$ is defined, we proceed to construct the Vietoris-Rips complex $VR(X, \epsilon)$ as follows (see, \eg~\cite{gromov1987hyperbolic, hatcher2002algebraic, munkres2018elements} for further details):
\begin{definition}[Vietoris-Rips Complex]
\label{Def:VRC}
Given a set of points $X = x_\alpha \subset \mathbb{R}^n$ in Euclidean $n$-space and a fixed radius $\epsilon$, the \textbf{Vietoris-Rips complex} of $X$, denoted $VR(X,\epsilon)$, is the abstract simplicial complex where each $k$-simplex corresponds to an unordered $(k + 1)$-tuple of points in X that are pairwise within Euclidean distance $\epsilon$ of each other.
\end{definition}

Here, an \textit{Abstract Simplicial Complex} (ASC) captures the topological structure of data by representing interactions among points through simplices. The definition of ASC (see, \eg ~\cite{joslyn2020sheaf, hatcher2002algebraic}) is provided in Definition \ref{def:asc}.

\begin{definition}[Abstract Simplicial Complex]
\label{def:asc}
An \textbf{abstract simplicial complex} $C$ is formed from a finite set $V_C$ and is defined as a collection of subsets of $V_C$ such that for every subset $\sigma \in C$, all subsets $\gamma \subseteq \sigma$ are also elements of $C$. Each subset $\sigma \in C$ is called a \textbf{simplex}. Any subset $\gamma$ of a simplex $\sigma$ is said to be a \textbf{face} of $\sigma$. Moreover, if $\sigma$ has $d+1$ elements, then it is called a \textbf{d-face} of $C$, where \textbf{d} denotes its dimension. \textbf{Vertices} are zero-dimensional faces (single elements of $V_C$), while \textbf{edges} are one-dimensional faces.
\end{definition}

Observe that in constructing the Vietoris-Rips complex, we include \( k \)-simplices based on whether the distances between all pairs of points in the simplex are within the specified threshold \( \epsilon \).

\subsubsection{Persistent Homology Computation}
The selection of the radius $\epsilon$ is a critical aspect of this process, as it determines the scale at which topological features are detected. This analysis tracks the "birth" and "death" of these features within a filtration—a nested sequence of simplicial complexes.  Choosing $\epsilon$ can be guided by analyzing the distribution of distances within $D$ or through persistent homology, which examines how the topology of the complex evolves as $\epsilon$ varies. In practice, \(\epsilon\) is typically chosen from a discrete set of values, \(\{\epsilon_i \mid i \in \mathbb{N}\}\), which correspond to critical thresholds in the topological structure derived from the distance matrix.
 Formally, persistent homology (see, \eg ~\cite{ghrist2008barcodes}) is defined as:

\begin{definition}[Persistent Homology]
\label{Def:PH} Let \( C = (C_i)_{i \geq 0} \) be a filtration of simplicial complexes with homology groups \( H_p(C_i) \) (over a field). For \( p \geq 0 \), the \textbf{\( p^{\text{th}} \)-persistent homology groups} of \( C \) are defined to be the images of the induced homomorphisms \( x_p : H_p(C_i) \to H_p(C_j) \) for all \( 0 \leq i \leq j \). We denote these groups by \( H^{p}_{i \to j}(C) \).
The \textbf{$p^{th}$-persistent Betti numbers} $\beta_p^{i,j}$ are the dimensions of these groups as vector spaces (over the given field).
\end{definition}

Persistent homology examines how topological features, such as holes, evolve as the scale varies. The lifespan of these features is visualized through persistence diagrams or barcodes, which summarize their persistence across scales. In persistent homology, a barcode represents the lifespan of topological features, similar to how (topological) Betti numbers (\(\beta_k\)) quantify the number of \( k \)-dimensional holes. Barcodes extend this concept by capturing the persistence of these features over varying parameters, filtering out noise and highlighting significant structures.

The topological features identified through persistent homology provide profound insights into data. In MTS analysis, holes can indicate cyclical behaviors or repeating patterns, such as economic cycles or seasonal variations in climate data. They also reveal feedback mechanisms and dependencies among variables, as seen in predator-prey models or financial market dynamics. In image analysis, holes and higher-dimensional features uncover significant shapes or patterns, offering a deeper understanding of the structure of data.

\subsection{Persistent Homology-based Feature Selection}
This approach utilizes persistent barcodes, focusing on the death times of all topological structures formed within the dataset. Empirical evidence shows that using the median death time across all structures provides an optimal Vietoris-Rips parameter $\epsilon$. This parameter enhances connectivity by effectively pruning variables that are not connected to any other variables, leading to improved performance. Algorithm \ref{alg:OptimizeVietorisRipsParameter} provides a step-by-step breakdown of the method.

\begin{algorithm}
\caption{Optimize Vietoris-Rips Parameter}
\label{alg:OptimizeVietorisRipsParameter}
\begin{algorithmic}[1] 

\REQUIRE Correlation matrix of variables in a MTS dataset $Corr_X$
\ENSURE Optimal Vietoris-Rips parameter $\epsilon$, pruned dataset with improved connectivity

\STATE Initialize an empty list \texttt{DeathTimes}
\STATE Compute distance matrix $D_X$ from the correlation matrix $Corr_X$
\STATE Compute persistent homology given the maximal 
$N$ such that $\beta_{N}\neq 0$.

\FOR{each $p$ in $\{1, \dots, N\}$}
    \STATE Extract the death times of topological structures from the computed persistent homology at $p$ (see Def. \ref{Def:PH})
    \STATE Append these death times to the \texttt{DeathTimes} list
\ENDFOR

\STATE Compute the median death time from the \texttt{DeathTimes} list and assign it to $\epsilon_{\text{optimal}}$

\STATE Construct the Vietoris-Rips complex on the dataset $X$ using $\epsilon_{\text{optimal}}$

\STATE Prune variables (nodes) not connected to any other variables in the Vietoris-Rips complex

\RETURN $\epsilon_{\text{optimal}}$ and the pruned dataset

\end{algorithmic}
\end{algorithm}

In this paper, we demonstrate that utilizing the pruned dataset allows us to maintain performance levels while eliminating, on average, $30\%$ of the variables in four tree-based and boosting classification models.

\section{Enhancing Explainability with Sheaf Structures While Preserving Model Performance}

Building on the optimal Vietoris-Rips parameter \(\epsilon_{\text{optimal}}\) and the pruned dataset, we enhance model explainability with sheaf structures. Sheaf theory maps interactions among the remaining variables, creating explanatory vectors that clarify each variable's contribution to model predictions.

\subsection{Sheaf Structures}
A sheaf systematically assigns data to parts of a space. Specifically, a \textit{sheaf of vector spaces} associates a vector space with each face and a linear map to every attachment (such that suitable compatibility conditions are satisfied), while a \textit{sheaf of groups} assigns groups to faces and group homomorphisms to attachments. However, a sheaf does more than just assign data to individual objects; it systematically ensures that these objects are interconnected in a controlled and coherent manner. This approach allows for seamless integration of local and global data interactions. The sheaf model describes how multivariate data is temporally aligned, as explained below. At any given time $ t $, each vertex in the model is assigned the latest reading, represented by a data point from its stalk space. The stalk space, in this context, is the collection of all possible data values that can be assigned to a point, capturing the local data structure. These readings are then transmitted to higher-dimensional faces for comparison using restriction maps. If two measurements on an edge are identical, then this shared value is assigned to the edge, and the process continues. When a complete assignment of values across the entire complex is possible, it is known as a global section. However, discrepancies may arise, necessitating the concept of an assignment. This framework leads to the following definition, as outlined, in particular, in~\cite{joslyn2020sheaf}.

\begin{definition}[Assignment and Global Section]
Let $\mathcal{S}$ be a sheaf on an abstract simplicial complex $X$. An \textbf{assignment} is a function $\gamma$ that maps each face $ x \in X $ to a value $\gamma(x) \in \mathcal{S}(x)$. A \textbf{partial assignment} is similarly defined, but it applies to a subset $ X' \subset X $: a function $\alpha$ that assigns a value $\alpha(x) \in \mathcal{S}(x)$ to each face $ x \in X'$. An assignment $ \gamma $ is called a \textbf{global section} if it satisfies the condition $\mathcal{S}(x \rightarrow  y)(\gamma(x)) = \gamma(y)$ for every inclusion $ x \rightarrow y$ of faces in $ X $.
\end{definition}

Consistency structures relax the constraints on data alignment, using boolean functions to evaluate the level of agreement between sheaf values over faces. A \textit{conventional consistency structure} $(X, \mathcal{S}, \mathbf{C})$ checks for exact matches, while an \textit{$\epsilon$-approximate consistency structure}, defined as a triple $(X, \mathcal{S}, \mathbf{C}_\epsilon)$, allows for slight discrepancies, measured by a consistency function $\delta$, defined as the root of the average trace of the data’s covariance matrix. Formally,~\cite{joslyn2020sheaf} specifically defines such structure as well as the \textit{consistency function} $\delta$ recursively as follows.
We choose an $\epsilon>0$. 
Then:
\begin{equation}
\textbf{C}_\epsilon(\beta) = 
\begin{cases}
1, \quad if \; \delta(\beta) \leq \epsilon \\
0, \quad otherwise.
\end{cases}
\end{equation}
Here, $\textbf{C}_\epsilon$ is a function on $d$-faces $\beta$ for $d>0$.
To define $\delta$ we consider $z_i = \mathcal{S}({v_i} \rightarrow \beta)(\gamma(v_i))$ for all vertices $v_i$ of $\beta$ and a given fixed assignment $\gamma$. For such a face $\beta$ we set $Y=[z_1, z_2, \dots, z_{d+1}]$ and consider its (empirical) covariance matrix $\Sigma_Y$. Then the consistency function $\delta$ evaluated on $\beta$ is defined as:
\begin{equation}
\label{eq:consistencymeasure}
\delta(\beta) = \sqrt{\frac{1}{|d+1|} \textbf{Tr}(\Sigma_Y)}\,.
\end{equation}

\textit{Pseudosections} extend the concept of global sections by ensuring that data assignments remain consistent up to a tolerance $\epsilon$, \ie $\textbf{C}_\epsilon(\beta)=1$ for all possible $\beta \in X$. The \textit{consistency radius} is the smallest $\epsilon$ for which a pseudosection exists, defined by the constraints of the data associated with the vertices.

The consistency structures of sheaf theory define \textit{maximally consistent subcomplexes}~\cite{joslyn2020sheaf}, which collectively cover the simplicial complex $X$ and ensure the preservation of the consistency in partial data assignments. In order to evaluate the consistency of multivariate data across varying $\epsilon$ values, we use \textit{consistency filtrations}. The filtration of covers corresponds to landmarks defined by increasing $\epsilon$ values, $\epsilon_0 = 0 < \epsilon_1 < \dots < \epsilon_{t-1} < \epsilon_{t} = \epsilon^*$, where $\epsilon^*$ is the consistency radius—the smallest \(\epsilon\) making the assignment $\gamma$ a $(X, \mathcal{S}, \mathbf{C}_{\epsilon})$-pseudosection. 

\subsection{Enhancing Multivariate Time Series Classification Explainability Using Sheaf Consistency Filtration}
As shown in Algorithm \ref{alg:OptimizeVietorisRipsParameter}, we can construct an ASC using the connections identified by applying the optimal parameter $\epsilon_{\text{optimal}}$ to the correlation matrix. We can utilize this structure in order to construct a sheaf model by assigning a data vector space to each vertex in the ASC and applying an identity function to each directed edge connecting a face to a higher-order face. Next, we apply the \(\epsilon\)-approximate consistency structure in order to obtain the consistency filtration for the variables in the MTS at each time step. This filtration is then integrated into the input as additional features alongside the original variables. Incorporating these extra dimensions allows the model to leverage enriched information, enabling us to analyze the impact of these additional features on the model's predictions.

\section{Integrating Persistent Homology and Sheaf Theory in \PHeatPruner}

This section demonstrates how \PHeatPruner leverages the combined strengths of persistent homology and sheaf theory to improve feature selection and model performance in Multivariate Time Series (MTS) classification.

\PHeatPruner begins by analyzing the correlation matrix of MTS data through persistent homology. This method transforms the correlation matrix into a distance matrix and constructs a Vietoris-Rips complex. By varying the threshold \( \epsilon \), persistent homology captures topological features at multiple scales. The optimal \( \epsilon \) is chosen to highlight and retain significant features, effectively pruning less critical variables. This process simplifies the dataset, reducing its dimensionality while preserving essential characteristics. Additionally, the construction of the Vietoris-Rips complex as an abstract simplicial complex (ASC) provides a foundation for further interpretability analysis by incorporating data vector spaces.

Following the pruning step, \PHeatPruner applies sheaf theory to enhance interpretability. Sheaf theory assigns vector spaces to the vertices in the ASC formed by the pruned variables and defines linear maps to represent their interactions. This structure captures both local and global relationships within the data. It generates explanatory vectors that elucidate each variable's contribution to the model's predictions. These explanatory vectors are then integrated into the machine learning model as additional features, offering deeper insights into the data’s structure and improving overall interpretability.

The integration of persistent homology and sheaf theory in \PHeatPruner creates an end-to-end unsupervised feature pruning framework. This pipeline not only enhances model performance by simplifying the data but also paves the way for deeper explainability. By reducing dimensionality and adding meaningful explanatory features, \PHeatPruner provides a robust and interpretable approach to MTS classification, facilitating informed decision-making across various applications.

\section{Experimental evaluation}
In this paper, we present two distinct experiments, which are described below:
\begin{enumerate}
\item First, we evaluate \PHeatPruner using the UAE Archive data, which comprises 30 different types of MTS. We examine the accuracy variations across four classical machine learning models: Random Forest, CatBoost, XGBoost, and LightGBM. Detailed information about each dataset is available in~\cite{bagnall2018uea}.
\item Second, we focus on a specific dataset from our project, aimed at the early detection of mastitis in dairy cows. For this dataset, whose details could be found in \cite{okashongwe2023resampling}, we assess the accuracy of the same four models and evaluate the interpretability by extracting the respective SHAP values from each model for every scenario.
\end{enumerate}

\subsection{Multivariate Time Series Classification Benchmark}
In this experiment, we tested $19$ out of $30$ datasets from the UEA Archive. We excluded eleven datasets for the following reasons: (1) Six datasets (Articulary Word Recognition, Atrial Fibrillation, Cricket, Libras, Pen Digits and SelfR Regulation SCP1) showed no change in the number of variables after applying \PHeatPruner, and (2) the other five datasets (Duck Duck Geese, Eigen Worms, Face Detection, Insect Wingbeat, and PEMS-SF) could not be processed by the machine learning models used in our analysis.

Table \ref{tab:ueaacc1} demonstrates that variable pruning reduces the feature set size while maintaining or improving model accuracy. We evaluated Random Forest (RF), CatBoost (CB), XGBoost (XGB), and LightGBM (LGBM) across various datasets before and after pruning, with each model initialized with 100 estimators and a random seed of 42. Post-pruning, accuracy often improves or remains stable. For example, RF and LGBM show increased accuracy in the "BasicMotions" dataset after a $16.67\%$ feature reduction, and multiple models show gains in the "Epilepsy" dataset with a $33.33\%$ reduction. However, the "FingerMovements" dataset experienced a decrease in accuracy, indicating the importance of specific features. On average, $30\%$ of variables were pruned, enhancing or maintaining model accuracy. \PHeatPruner notably improves baseline models; for instance, RF's accuracy in "BasicMotions" rose from 0.90 to 0.95-1.00, and XGB's in "Motor Imagery" increased from 0.44 to 0.58 after pruning 12 variables, nearing ShapeNet's $0.61$. In "SelfRegulationSCP2," CB's accuracy increased from $0.48$ to $0.51$, surpassing WEASEL+MUSE and aligning with state-of-the-art models. While XCM is a robust deep learning model that outperforms other methods on most UEA Archive datasets, \PHeatPruner enables classical models to exceed XCM's performance on specific datasets like "Hand Movement Direction" and "Motor Imagery".

	\begin{table}[h!]
    \caption{Evaluation of Model Accuracy for Random Forest (RF), CatBoost (CB), XGBoost (XGB), and LightGBM (LGBM) Before (values in brackets) and After Variable Pruning for Benchmark Datasets. The results are color-coded as: \color{teal}Green-Win \color{black}(improvement), \color{olive}Yellow-Deuce \color{black}(no change), and \color{red}Red-Lose \color{black}(decline). Abbreviations for datasets: BM (BasicMotions), CT (CharacterTrajectories), EP (Epilepsy), EC (EthanolConcentration), ER (ERing), FM (FingerMovements), HMD (HandMovementDirection), HW (Handwriting), HB (Heartbeat), JW (JapaneseVowels), LSST (LSST), MI (MotorImagery), NATOPS (NATOPS), PS (Phoneme), RP (RacketSports), SCP2 (SelfRegulationSCP2), SAD (SpokenArabicDigits), SWJ (StandWalkJump), UW (UWaveGestureLibrary).}
    \centering
    \small
    \label{tab:ueaacc1}
\scalebox{0.825}{
    \begin{tabular}{||c@{\hskip 0.1in}c@{\hskip 0.1in}c@{\hskip 0.1in}c@{\hskip 0.1in}c@{\hskip 0.1in}c@{\hskip 0.1in}c@{\hskip 0.1in}c||@{\hskip 0.1in}c@{\hskip 0.1in}c@{\hskip 0.1in}c||} 
    \hline
    {\textbf{Dataset}} & \textbf{RF} &  {\textbf{CB}} & {\textbf{XGB}} & \textbf{LGBM} &  \textbf{Orig.} & \textbf{After} & {\textbf{\%}}  & \textbf{Shape} & \textbf{WEASEL} & \textbf{XCM}  \\ [0.5ex] 
    &  & & &  & \textbf{\#Var.} &\textbf{\#Var.} & \textbf{Pruned} & \textbf{Net} & \textbf{+MUSE} &   \\ [0.7ex] 
    \hline\hline
    BM & \color{teal}{0.95}\color{black}(0.9) & \color{red}{0.78}\color{black}(0.82) & \color{olive}{0.82}\color{black}(0.82) & \color{red}{0.78}\color{black}(0.8) & 6 & 5 & 16.67 & \textbf{1.0} & \textbf{1.0} & \textbf{1.0}\\ 
    \hline
    CT & \color{olive}{0.98}\color{black}(0.98) & \color{teal}{0.97}\color{black}(0.96) & \color{red}{0.96}\color{black}(0.97) & \color{olive}{0.97}\color{black}(0.97) & 3 & 2 & 33.33 & 0.98 & 0.99 & \textbf{0.995}\\     
    \hline
    EP & \color{teal}{0.82}\color{black}(0.8) & \color{red}{0.79}\color{black}(0.83) & \color{teal}{0.72}\color{black}(0.7) & \color{red}{0.75}\color{black}(0.77) & 3 & 2 & 33.33 & 0.987 & \textbf{0.993} & \textbf{0.993} \\ 
    \hline
    EC & \color{teal}{0.42}\color{black}(0.41) & \color{teal}{\textbf{0.45}}\color{black}(0.44) & \color{olive}{0.43}\color{black}(0.43) & \color{olive}{0.43}\color{black}(0.43) & 3 & 2 & 33.33 & 0.312 & 0.316 & 0.346 \\  
    \hline
    ER & \color{red}{0.86}\color{black}(\textbf{0.91}) & \color{red}{0.80}\color{black}(0.84) & \color{teal}{0.63}\color{black}(0.6) & \color{olive}{0.17}\color{black}(0.17) & 3 & 2 & 33.33 & 0.133 & 0.133 & 0.133\\  
    \hline
    FM & \color{teal}{0.57}\color{black}(0.52) & \color{red}{0.53}\color{black}(0.57) & \color{red}{0.53}\color{black}(0.55) & \color{red}{0.54}\color{black}(0.57) & 28 & 16 & 42.8 & 0.58 & 0.54 & \textbf{0.6} \\  
    \hline
    HMD & \color{teal}{\textbf{0.53}}\color{black}(0.51) & \color{teal}{0.39}\color{black}(0.35) & \color{teal}{0.47}\color{black}(0.43) & \color{olive}{0.41}\color{black}(0.41) & 10 & 7 & 30 & 0.338 & 0.378 & 0.446\\ 
    \hline
    HW & \color{red}{0.16}\color{black}(0.18) & \color{olive}{0.17}\color{black}(0.17) & \color{red}{0.14}\color{black}(0.15) & \color{red}{0.15}\color{black}(0.17) & 3 & 2 & 33.33 & 0.451 &\textbf{0.549} & 0.412 \\ 
    \hline
    HB & \color{teal}{0.73}\color{black}(0.7) & \color{olive}{0.77}\color{black}(0.77) & \color{teal}{0.76}\color{black}(0.7) & \color{olive}{0.73}\color{black}(0.73) & 61 & 39 & 36 & 0.756 & 0.722 & \textbf{0.776} \\  
    \hline
    JW & \color{olive}{0.95}\color{black}(0.95) & \color{olive}{0.97}\color{black}(0.97) & \color{teal}{0.93}\color{black}(0.92) & \color{olive}{0.95}\color{black}(0.95) & 12 & 10 & 16.67 & 0.984 & 0.951 & \textbf{0.986}\\ 
    \hline
    LSST & \color{red}{0.52}\color{black}(0.54) & \color{teal}{0.47}\color{black}(0.46) & \color{red}{0.52}\color{black}(0.55) & \color{red}{0.54}\color{black}(0.55) & 6 & 4 & 33.33 & 0.59 & 0.315 & \textbf{0.612}\\  
    \hline
    MI & \color{teal}{0.54}\color{black}(0.52) & \color{teal}{0.55}\color{black}(0.49) & \color{teal}{0.58}\color{black}(0.44) & \color{red}{0.52}\color{black}(0.54) & 64 & 52 & 18.75 &\textbf{ 0.61} & 0.5 & 0.54  \\ 
    \hline
    NATOPS & \color{teal}{0.85}\color{black}(0.84) & \color{red}{0.82}\color{black}(0.84) & \color{teal}{0.85}\color{black}(0.83) & \color{red}{0.80}\color{black}(0.84) & 24 & 23 & 4.2 & 0.883 & 0.883 & \textbf{0.978} \\ 
    \hline
    PS & \color{olive}{0.11}\color{black}(0.11) & \color{olive}{0.09}\color{black}(0.09) & \color{olive}{0.1}\color{black}(0.1) & \color{olive}{0.1}\color{black}(0.1) & 11 & 6 & 45.45 & 0.298 & 0.026 & \textbf{0.225}\\ 
    \hline
    RS & \color{teal}{0.86}\color{black}(0.85) & \color{teal}{0.84}\color{black}(0.83) & \color{red}{0.79}\color{black}(0.82) & \color{red}{0.80}\color{black}(0.82) & 6 & 4 & 33.33 & 0.882 & 0.829 & \textbf{0.895}\\  
    \hline
    SCP2 & \color{red}{0.49}\color{black}(0.49) & \color{teal}{0.51}\color{black}(0.48) & \color{teal}{0.51}\color{black}(0.48) & \color{red}{0.53}\color{black}(0.54) & 7 & 6 & 14.28 & \textbf{0.578} & 0.5 & 0.544 \\ 
    \hline
    SAD & \color{olive}{0.95}\color{black}(0.95) & \color{olive}{0.95}\color{black}(0.95) & \color{red}{0.95}\color{black}(0.96) & \color{red}{0.95}\color{black}(0.96) & 13 & 10 & 23.07 & 0.975 & 0.986 & \textbf{0.995} \\ 
    \hline
    SWJ & \color{red}{0.4}\color{black}(0.53) & \color{olive}{\textbf{0.53}}\color{black}(\textbf{0.53}) & \color{olive}{0.47}\color{black}(0.47) & \color{olive}{0.33}\color{black}(0.33) & 4 & 3 & 25 & \textbf{0.533} & \textbf{0.533} & 0.4\\ 
    \hline
    UW & \color{red}{0.81}\color{black}(0.84) & \color{red}{0.75}\color{black}(0.81) & \color{red}{0.7}\color{black}(0.73) & \color{red}{0.75}\color{black}(0.79) & 3 & 2 & 33.33 &\textbf{ 0.906} & 0.812 & 0.894  \\  
    \hline
    \end{tabular}
}
    \end{table}

\subsection{Evaluation of Prediction Performance and Interpretability for Early Mastitis Detection in Dairy Cows Dataset}
The dataset used in the second experiment contains time series data of various physiological and environmental factors for dairy cows in a barn. These factors include, in particular:
\begin{inparaenum}[(1)]
    \item \textit{Rumination Index}: Time spent chewing cud, indicating digestive health.
    \item \textit{Raw Temperature}: Unadjusted temperature readings of the cows.
    \item \textit{Adjusted Temperature}: Temperatures normalized to $39-40^\circ$ Celsius for consistent comparisons.
    \item \textit{Temperature Excluding Drinking}: Temperature readings that exclude fluctuations from drinking water.
    \item \textit{Milk Index}: Metrics related to milk production, reflecting dairy productivity and cow health.
\end{inparaenum}

Besides, some engineered features, specifically \textit{quarter hour measures}, \textit{half hour measures}, \textit{hourly measures}, \textit{3 hour measures}, \textit{12 hour measures} are also included. These factors are recorded over time and organized into $72,480$ sequences of four consecutive days. Each sequence is labeled to indicate the presence or absence of mastitis, with $930$ sequences identified as leading to mastitis.

\subsubsection{Evaluation of the Prediction Performance of the Pruning method}
\begin{figure}
\includegraphics[width=\textwidth]{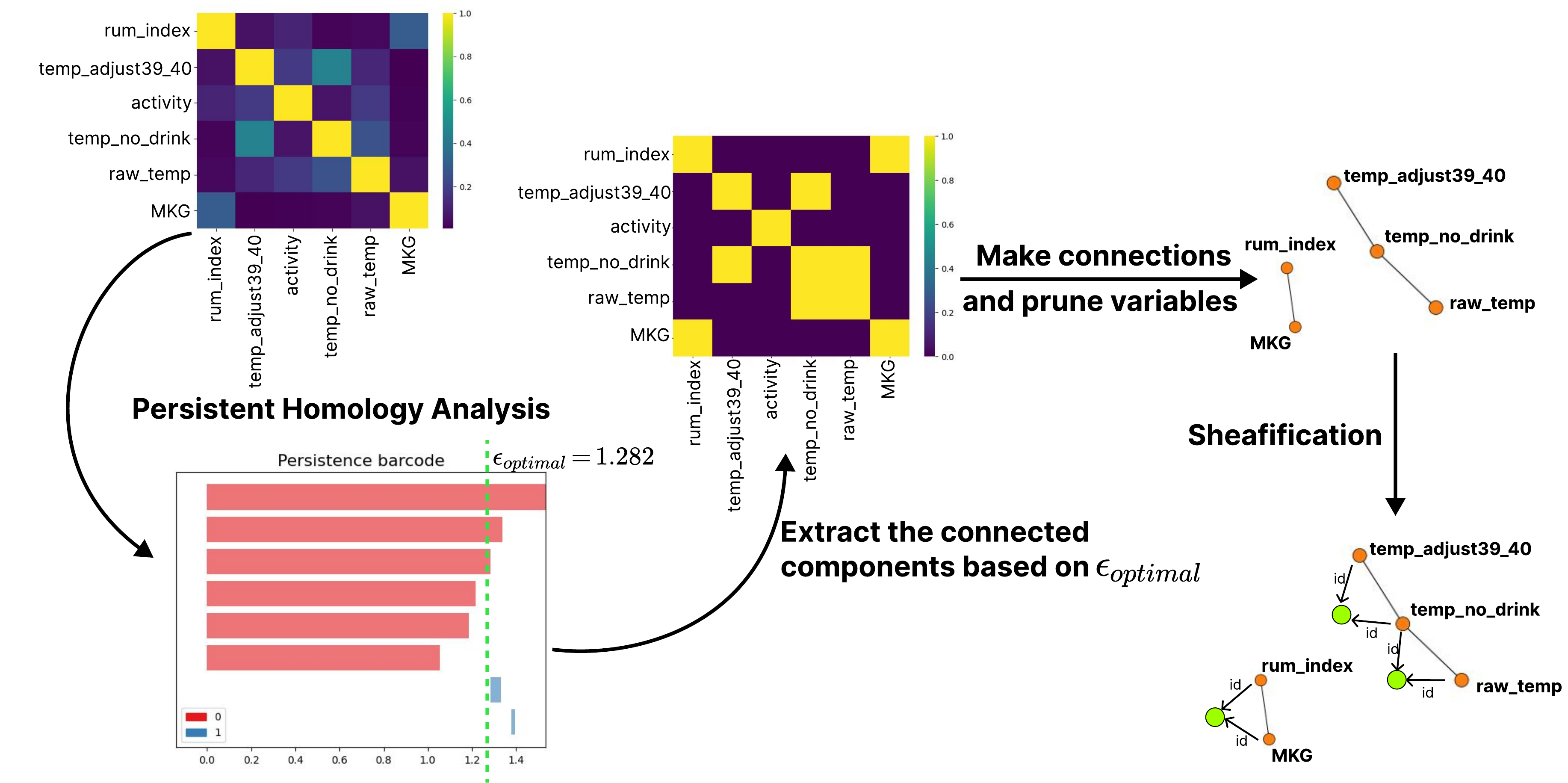}
\caption{Persistent Homology-Based Feature Selection and Sheafification for Mastitis Detection.} \label{fig:mastitis}
\end{figure}

In order to facilitate variable pruning, we first construct a correlation matrix. Using this matrix, Algorithm \ref{alg:OptimizeVietorisRipsParameter} determines the optimal distance parameter \( \epsilon_{\text{optimal}} \), based on the median "death times" of the topological features. Variables with no connections are pruned, resulting in a reduced dataset. For instance, the feature "Activity" is eliminated by this process, as shown in Fig. \ref{fig:mastitis}.

The experiment uses hyperparameter-tuned classical machine learning models with the following configurations:
\begin{inparaenum}[(1)]
    \item \textbf{Random Forest}: \texttt{n\_estimators} = 100.
    \item \textbf{CatBoost}: \texttt{depth} = 10, \texttt{iterations} = 440, \texttt{l2\_leaf\_reg} = 1.077,\texttt{learning\_rate} = 0.0722.
    \item \textbf{XGBoost}: \texttt{colsample\_bytree} = 0.637, \texttt{learning\_rate} = 0.0971, \texttt{max\_depth} = 6, \texttt{n\_estimators} = 305, \texttt{reg\_alpha} = 0.154, \texttt{reg\_lambda} = 0.659,\texttt{subsample} = 0.901.
    \item \textbf{LightGBM}: \texttt{colsample\_bytree} = 0.663, \texttt{learning\_rate} = 0.0522, \texttt{max\_depth} = 15, \texttt{min\_child\_samples} = 30, \texttt{n\_estimators} = 383,\texttt{num\_leaves} = 41, \texttt{subsample} = 0.692.
\end{inparaenum}

The results are summarized in Table \ref{tab:cowacc} as the average result of 5 fold cross-validation throughout the dataset. Overall, Variable pruning generally improves model performance, notably enhancing recall in Random Forest (RF), which is vital for managing the dataset's class imbalance. LightGBM (LGBM) maintains its high performance, indicating prior optimization. Crucially, no metrics decline post-pruning, affirming the pruning's effectiveness.

    \begin{table}[h!]
    \caption{Evaluation of Model Accuracy for Random Forest (RF), CatBoost (CB), XGBoost (XGB), and LightGBM (LGBM) Before and After Variable ($1/6 \equiv 16.67\%$ of the variables have been pruned) Pruning for Mastitis Dairy Cow Dataset. The colors are marked as: \color{teal}Green-Win, \color{olive}Yellow-Deuce and \color{red}Red-Lose.}
    \centering
    \small
    \label{tab:cowacc}
    \begin{tabular}{||c@{\hskip 0.1in}c@{\hskip 0.1in}c@{\hskip 0.1in}c@{\hskip 0.1in}c||} 
    \hline
    {\textbf{Models}} & \textbf{Accuracy} &  {\textbf{Precision}} & {\textbf{Recall}} & \textbf{F1 Score} \\ [0.5ex] 
   
    \hline\hline
    RF &  0.996 & 1.0 & 0.97 & 0.98 \\ 
    \hline
    CB &  1.0 & 1.0 & 0.996 & 0.998  \\    
    \hline
    XGB &  1.0 & 1.0 & 0.998 & 1.0  \\  
    \hline
    LGBM&  1.0 & 1.0 & {1.0}& 1.0 \\  
    \hline \hline
    RF After Pruned &  \color{teal}1.0 & \color{olive}1.0 & \color{teal}0.99& \color{teal}1.0  \\  
    \hline
    CB After Pruned &  \color{olive}1.0 & \color{olive}{1.0} &\color{teal}0.998& \color{olive}0.998  \\ 
    \hline
    XGB After Pruned &  \color{olive}1.0 & \color{olive}1.0 & \color{teal}{1.0}& \color{olive}1.0  \\ 
    \hline
    LGBM After Pruned &  \color{olive}{1.0} & \color{olive}1.0 &\color{olive}1.0& \color{olive}1.0 \\  
    \hline
    \end{tabular}
    \end{table}

\subsubsection{Evaluating the Interpretability of the Sheaf Structure on the Pruned Dataset’s Induced Simplicial Complex}
From Figure~\ref{fig:mastitis}, the pruned dataset's simplicial complex contains 1-faces (edges) as the highest structures. The edges link the pairs: \{"Milk Index", "Rumination Index"\}, \{"Adjusted Temperature", "Temperature Excluding Drinking"\}, and \{"Raw Temperature", "Temperature Excluding Drinking"\}. We introduce a sheaf structure to this complex to compute the consistency filtration for each pair using Eq. \ref{eq:consistencymeasure}. These consistency values are added as features in the models. We then evaluate the impact on model performance and interpretability using SHAP values~\cite{lundberg2017shap}. 

\begin{figure}[h!]
\includegraphics[width=0.7\textwidth]{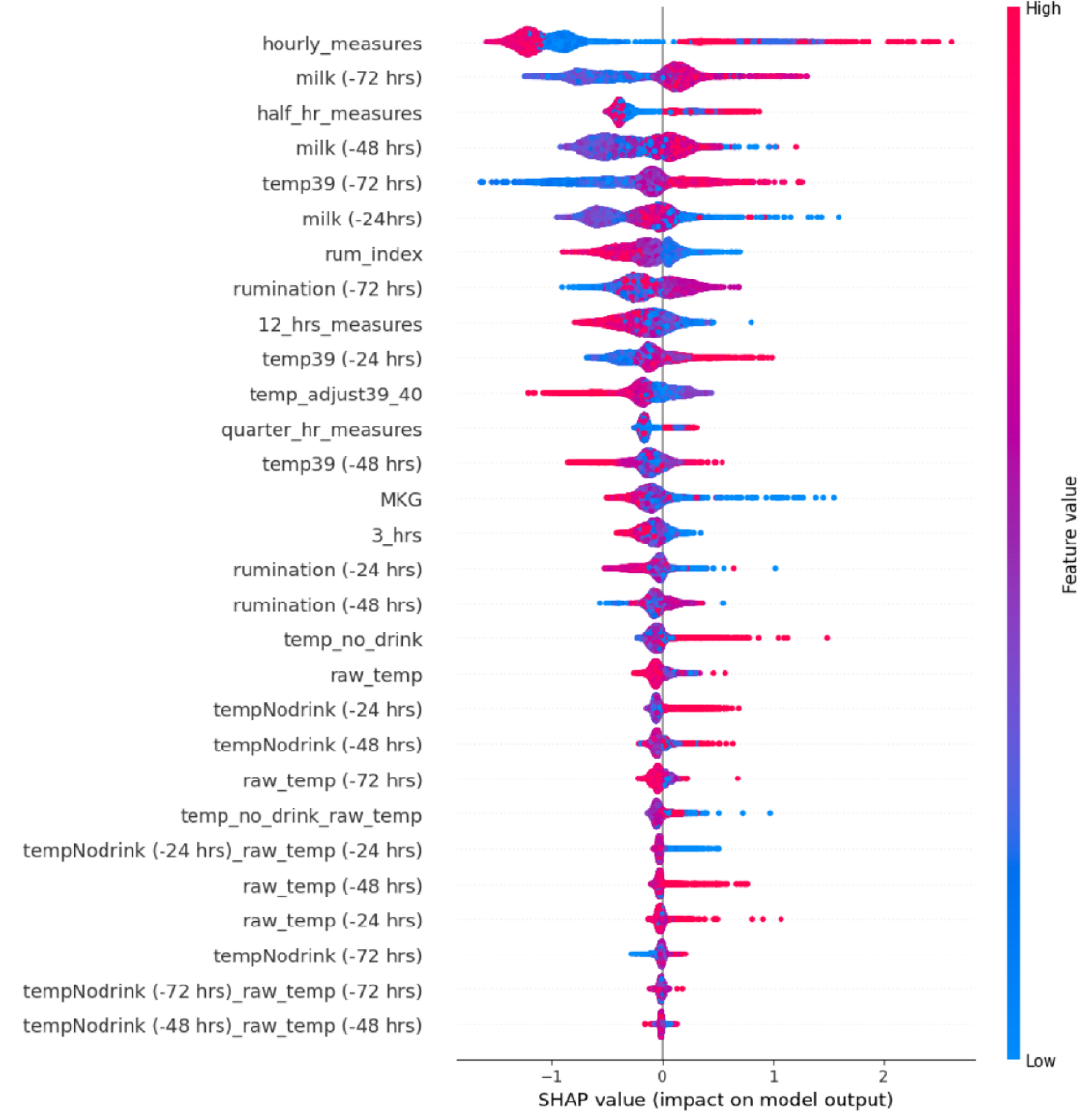}
\centering
\caption{SHAP Value Analysis of Input Features on XGBoost Model Outcomes for the Mastitis Class, illustrating the SHAP values for various input features and their impact on the XGBoost model’s predictions for mastitis. Features are color-coded, with red indicating higher values and blue indicating lower values. The x-axis represents the magnitude of each feature's impact on the model's output, showing how changes in feature values influence the model's prediction for the mastitis class.} \label{fig:xgbSHAP}
\end{figure}

Adding extra features to the pruned dataset does not alter the performance of the machine learning models compared to using the pruned dataset alone, so these results are not reported separately. This consistency confirms the reliability of the interpretations provided by the additional features. Since all models performed similarly, we focus on the best-performing model, XGBoost, to analyze the impact of these features on model outcomes and their explainability. Figure \ref{fig:xgbSHAP} shows that while three simplices were added as features, only one significantly contributes to predicting mastitis, as highlighted by the SHAP value graph.

The results of the SHAP analysis highlights the critical role of the variables \\"temp\_no\_drink\_raw\_temp" in the XGBoost model for predicting mastitis in dairy cows, especially with time lags. Although these variables show lower overall feature importance, their impact grows over different intervals. Specifically, "Raw Temperature" and "Temperature Excluding Drinking" correlate with increased mastitis risk $24$ to $48$ hours before onset. As the gap between these temperatures narrows in the $24$ hours before sickness, the probability of mastitis rises. This dynamic suggests that both individual values and their convergence are key for early mastitis detection.

\section{Conclusion}
This paper combines persistent homology and sheaf theory to simplify MTS  classification and enhance interpretability. In \PHeatPruner, persistent homology prunes up to $45\%$ of variables, maintaining or improving accuracy in models like Random Forest, CatBoost, XGBoost, and LightGBM. Sheaf theory adds explanatory vectors, providing deeper insights. Our experiments demonstrate that this method preserves model accuracy and highlights the importance of specific features, such as the narrowing gap between "Raw Temperature" and "Temperature Excluding Drinking" for early mastitis detection in dairy cows. Future work will extend this approach to multimodal datasets and incorporate neuro-symbolic methods to further enhance model accuracy and interpretability, also enabling more informed decision-making.

\begin{credits}
\subsubsection{\ackname} This research was supported by the Lower Saxony Ministry of Science and Culture (MWK), funded  through the zukunft.niedersachsen program of the Volkswagen Foundation. The data in this research was collected, processed, and evaluated with funding from the MEDICow project, supported by the German Federal Ministry of Food and Agriculture (BMEL) under the "Livestock Husbandry" program (grant number: 28N-2-066-01, project MEDICow). We extend our gratitude to Dr. Tina Kabelitz for her invaluable support in data collection. We also thank Prof. Thomas Amon and Prof. Barbara Amon for their assistance with facilities and organizational support.
\end{credits}

%
%
%
\bibliographystyle{splncs04}
\bibliography{ref}

\begin{thebibliography}{10}
\providecommand{\url}[1]{\texttt{#1}}
\providecommand{\urlprefix}{URL }
\providecommand{\doi}[1]{https://doi.org/#1}

\bibitem{abramsky2015contextuality}
Abramsky, S., Barbosa, R.S., Kishida, K., Lal, R., Mansfield, S.: Contextuality, cohomology and paradox. arXiv preprint arXiv:1502.03097  (2015)

\bibitem{bagnall2018uea}
Bagnall, A., Dau, H.A., Lines, J., Flynn, M., Large, J., Bostrom, A., Southam, P., Keogh, E.: The uea multivariate time series classification archive, 2018. arXiv preprint arXiv:1811.00075  (2018)

\bibitem{berwald2013automatic}
Berwald, J., Gidea, M., Vejdemo-Johansson, M.: Automatic recognition and tagging of topologically different regimes in dynamical systems. arXiv preprint arXiv:1312.2482  (2013)

\bibitem{bodnar2023topological}
Bodnar, C.: Topological deep learning: graphs, complexes, sheaves. Ph.D. thesis (2023)

\bibitem{bredon2012sheaf}
Bredon, G.E.: Sheaf theory, vol.~170. Springer Science \& Business Media (2012)

\bibitem{carlsson2021topological}
Carlsson, G., Vejdemo-Johansson, M.: Topological data analysis with applications. Cambridge University Press (2021)

\bibitem{chazal2021introduction}
Chazal, F., Michel, B.: An introduction to topological data analysis: fundamental and practical aspects for data scientists. Frontiers in artificial intelligence  \textbf{4}, ~108 (2021)

\bibitem{choi2016retain}
Choi, E., Bahadori, M.T., Sun, J., Kulas, J., Schuetz, A., Stewart, W.: Retain: An interpretable predictive model for healthcare using reverse time attention mechanism. Advances in neural information processing systems  \textbf{29} (2016)

\bibitem{curry2014sheaves}
Curry, J.M.: Sheaves, cosheaves and applications. University of Pennsylvania (2014)

\bibitem{dequeant2008comparison}
Dequ{\'e}ant, M., Ahnert, S., Edelsbrunner, H., Fink, T., Glynn, E., et~al.: Comparison of pattern detection methods in microarray time series of the  (2008)

\bibitem{edelsbrunner2014short}
Edelsbrunner, H.: A short course in computational geometry and topology. No. Mathematical methods, Springer (2014)

\bibitem{fauvel2021xcm}
Fauvel, K., Lin, T., Masson, V., Fromont, {\'E}., Termier, A.: Xcm: An explainable convolutional neural network for multivariate time series classification. Mathematics  \textbf{9}(23), ~3137 (2021). \doi{https://doi.org/10.3390/math9233137}

\bibitem{gangopadhyay2021spatiotemporal}
Gangopadhyay, T., Tan, S.Y., Jiang, Z., Meng, R., Sarkar, S.: Spatiotemporal attention for multivariate time series prediction and interpretation. In: Proc. IEEE International Conference on Acoustics, Speech and Signal Processing (ICASSP). pp. 3560--3564. IEEE (2021)

\bibitem{ghrist2008barcodes}
Ghrist, R.: Barcodes: the persistent topology of data. Bulletin of the American Mathematical Society  \textbf{45}(1),  61--75 (2008)

\bibitem{ghrist2011network}
Ghrist, R., Hiraoka, Y.: Network codings and sheaf cohomology. IEICE Proceedings Series  \textbf{45}(A4L-C3) (2011)

\bibitem{gromov1987hyperbolic}
Gromov, M.: Hyperbolic groups. In: Gersten, S.M. (ed.) Essays in Group Theory, Mathematical Sciences Research Institute Publications, vol.~8, pp. 75--263. Springer-Verlag, New York (1987). \doi{10.1007/978-1-4613-9586-7_3}

\bibitem{hansen2019toward}
Hansen, J., Ghrist, R.: Toward a spectral theory of cellular sheaves. Journal of Applied and Computational Topology  \textbf{3}(4),  315--358 (2019)

\bibitem{hatcher2002algebraic}
Hatcher, A.: Algebraic Topology. Cambridge University Press, Cambridge (2002), \url{https://pi.math.cornell.edu/~hatcher/AT/ATpage.html}

\bibitem{joslyn2020sheaf}
Joslyn, C.A., Charles, L., DePerno, C., Gould, N., Nowak, K., Praggastis, B., Purvine, E., Robinson, M., Strules, J., Whitney, P.: A sheaf theoretical approach to uncertainty quantification of heterogeneous geolocation information. Sensors  \textbf{20}(12), ~3418 (2020)

\bibitem{joslyn2014towards}
Joslyn, C.A., Hogan, E., Robinson, M.: Towards a topological framework for integrating semantic information sources. In: STIDS. pp. 93--96 (2014)

\bibitem{karim2019multivariate}
Karim, F., Majumdar, S., Darabi, H., Harford, S.: Multivariate lstm-fcns for time series classification. Neural networks  \textbf{116},  237--245 (2019)

\bibitem{karlsson2016generalized}
Karlsson, I., Papapetrou, P., Bostr{\"o}m, H.: Generalized random shapelet forests. Data mining and knowledge discovery  \textbf{30},  1053--1085 (2016)

\bibitem{li2021shapenet}
Li, G., Choi, B., Xu, J., Bhowmick, S.S., Chun, K.P., Wong, G.L.H.: Shapenet: A shapelet-neural network approach for multivariate time series classification. In: Proceedings of the AAAI conference on artificial intelligence. vol.~35, pp. 8375--8383 (2021)

\bibitem{liang2018geoman}
Liang, Y., Ke, S., Zhang, J., Yi, X., Zheng, Y.: Geoman: Multi-level attention networks for geo-sensory time series prediction. In: IJCAI. vol.~2018, pp. 3428--3434 (2018)

\bibitem{liu2020dstp}
Liu, Y., Gong, C., Yang, L., Chen, Y.: Dstp-rnn: A dual-stage two-phase attention-based recurrent neural network for long-term and multivariate time series prediction. Expert Systems with Applications  \textbf{143},  113082 (2020). \doi{https://doi.org/10.1016/j.eswa.2019.113082}

\bibitem{lundberg2017shap}
Lundberg, S.M., Lee, S.I.: A unified approach to interpreting model predictions. In: Guyon, I., Luxburg, U.V., Bengio, S., Wallach, H., Fergus, R., Vishwanathan, S., Garnett, R. (eds.) Advances in Neural Information Processing Systems 30, pp. 4765--4774. Curran Associates, Inc. (2017)

\bibitem{malcolm2009sheaves}
Malcolm, G.: Sheaves, objects, and distributed systems. Electronic Notes in Theoretical Computer Science  \textbf{225},  3--19 (2009)

\bibitem{munkres2018elements}
Munkres, J.R.: Elements of algebraic topology. CRC press (2018)

\bibitem{okashongwe2023resampling}
Okashongwe: Resampling mastitis dataset (2023), \url{https://github.com/Okashongwe/ -resampling\_mastitis}, accessed: 2024-06-22

\bibitem{perea2015sw1pers}
Perea, J.A., Deckard, A., Haase, S.B., Harer, J.: Sw1pers: Sliding windows and 1-persistence scoring; discovering periodicity in gene expression time series data. BMC bioinformatics  \textbf{16},  1--12 (2015)

\bibitem{perea2015sliding}
Perea, J.A., Harer, J.: Sliding windows and persistence: An application of topological methods to signal analysis. Foundations of Computational Mathematics  \textbf{15},  799--838 (2015)

\bibitem{pham2023tsem}
Pham, A.D., Kuestenmacher, A., Ploeger, P.G.: Tsem: Temporally-weighted spatiotemporal explainable neural network for multivariate time series. In: Future of Information and Communication Conference. pp. 183--204. Springer (2023)

\bibitem{pham2023harnessing}
Pham, A.D., Le, A.D., Le, C.D., Pham, H.V., Vo, H.B.: Harnessing sheaf theory for enhanced air quality monitoring: Overcoming conventional limitations with topology-inspired self-correcting algorithm. In: Proceedings of the Future Technologies Conference. pp. 102--122. Springer (2023)

\bibitem{purvine2018topological}
Purvine, E., Aksoy, S., Joslyn, C., Nowak, K., Praggastis, B., Robinson, M.: A topological approach to representational data models. In: Human Interface and the Management of Information. Interaction, Visualization, and Analytics: 20th International Conference, HIMI 2018, Held as Part of HCI International 2018, Las Vegas, NV, USA, July 15-20, 2018, Proceedings, Part I 20. pp. 90--109. Springer (2018)

\bibitem{qin2017dual}
Qin, Y., Song, D., Chen, H., Cheng, W., Jiang, G., Cottrell, G.: A dual-stage attention-based recurrent neural network for time series prediction. arXiv preprint arXiv:1704.02971  (2017)

\bibitem{robinson2017sheaves}
Robinson, M.: Sheaves are the canonical data structure for sensor integration. Information Fusion  \textbf{36},  208--224 (2017)

\bibitem{robinson2020assignments}
Robinson, M.: Assignments to sheaves of pseudometric spaces. Compositionality  \textbf{2} (2020)

\bibitem{rosiak2022sheaf}
Rosiak, D.: Sheaf theory through examples. MIT Press (2022)

\bibitem{santoro2023higher}
Santoro, A., Battiston, F., Petri, G., Amico, E.: Higher-order organization of multivariate time series. Nature Physics  \textbf{19}(2),  221--229 (2023)

\bibitem{schafer2017multivariate}
Sch{\"a}fer, P., Leser, U.: Multivariate time series classification with weasel+ muse. arXiv preprint arXiv:1711.11343  (2017)

\bibitem{shokoohi2017generalizing}
Shokoohi-Yekta, M., Hu, B., Jin, H., Wang, J., Keogh, E.: Generalizing dtw to the multi-dimensional case requires an adaptive approach. Data mining and knowledge discovery  \textbf{31},  1--31 (2017)

\bibitem{theng2024feature}
Theng, D., Bhoyar, K.K.: Feature selection techniques for machine learning: a survey of more than two decades of research. Knowledge and Information Systems  \textbf{66}(3),  1575--1637 (2024)

\bibitem{wasserman2018topological}
Wasserman, L.: Topological data analysis. Annual Review of Statistics and Its Application  \textbf{5},  501--532 (2018)

\bibitem{wistuba2015ultra}
Wistuba, M., Grabocka, J., Schmidt-Thieme, L.: Ultra-fast shapelets for time series classification. arXiv preprint arXiv:1503.05018  (2015)

\end{thebibliography}
%




\end{document}